\documentclass{article} 
\usepackage{iclr2024_conference,times}

\usepackage{amsmath,amsfonts,bm}

\def\eqref#1{equation~\ref{#1}}

\def\1{\bm{1}}

\DeclareMathAlphabet{\mathsfit}{\encodingdefault}{\sfdefault}{m}{sl}
\SetMathAlphabet{\mathsfit}{bold}{\encodingdefault}{\sfdefault}{bx}{n}

\usepackage[utf8]{inputenc}
\usepackage{csquotes}
\usepackage{multirow} 
\usepackage{hyperref}
\usepackage{url}

\usepackage{microtype}
\usepackage{inconsolata}

\usepackage[title]{appendix}

\usepackage{times}
\usepackage{latexsym}
\usepackage{hyperref}
\usepackage{mathtools}
\usepackage{booktabs}

\usepackage{amssymb}
\usepackage{amsmath, bm}
\usepackage{float}
\usepackage{comment}

 \usepackage[ruled,vlined,linesnumbered]{algorithm2e}
\usepackage[T1]{fontenc}

\title{Apollo: Zero-shot MultiModal Reasoning with Multiple Experts}

\author{Daniela Ben-David \\
Faculty of Electrical and Computer Engineering, Technion, Israel\\
\texttt{bdaniela@campus.technion.ac.il} \\
\And
Tzuf Paz-Argaman\\
Department of Computer Science, Bar Ilan University, Israel\\
\texttt{tzuf.paz-argaman@biu.ac.il} \\
\AND
Reut Tsarfaty\\
Department of Computer Science, Bar Ilan University, Israel\\
\texttt{reut.tsarfaty@biu.ac.il}\\
}

\begin{document}

\maketitle

\begin{abstract}
We propose a modular framework that leverages the expertise of different foundation models over different modalities and domains in order to perform a single, complex, multi-modal task, without relying on prompt engineering or otherwise tailor-made multi-modal training. Our approach enables decentralized command execution and allows each model to both contribute and benefit from the expertise of the other models. Our method can be extended to a variety of foundation models (including audio and vision), above and beyond only language models, as it does not depend on prompts. We demonstrate our approach on two tasks. On the well-known task of stylized image captioning, our experiments show that our approach outperforms semi-supervised state-of-the-art models, while being zero-shot and avoiding costly training, data collection, and prompt engineering. We further demonstrate this method on a novel task, audio-aware image captioning, in which an image and audio are given and the task is to generate text that describes the image within the context of the provided audio. Our code is available at: 
\url{https://github.com/danielabd/Apollo-Cap}
\end{abstract}

\section{Introduction}

Humans perceive the world through different types of data (e.g., images and sounds) that they get from their senses. Similarly, to understand the world, artificial intelligence research also tries to solve problems that use multimodal data \citep{antol2015vqa, paz2020zest, ji2022abstract, rassin2023linguistic}. Solving multimodal tasks requires interpreting and reasoning over heterogeneous data, which poses several challenges, such as the training process \citep{Wang_2020_CVPR}.

Large pre-trained \textit{foundation} models demonstrate distinct expertise and encompass comprehensive knowledge within specific domains and modalities they are trained on. For example, BERT \citep{devlin2018bert} and GPT3 \citep{brown2020language}are proficient in processing language, while CLIP \citep{CLIP:radford2021learning} excels in grounding text to visual content. However, the large and increasing variety of multimodal tasks  (e.g., vision and language navigation \citep{ku2020room}, and video question-answering \citep{lei2018tvqa}), do not have \textit{foundation} models. Previous efforts to tackle complex multimodal tasks are either (1) fully-supervised, require expensive paired input and output task-specific data \citep{chen2019touchdown, Li_2022_CVPR}; 
(2) semi-supervised -- task-specific uncoupled data for each modality or domain \citep{capdec:nukrai2022text, Guo2019MSCapMI, MemCap:DBLP:conf/aaai/ZhaoWZ20, gan2017stylenet,su2022language}; (3) few-shot -- a few coupled task-specific examples; and (4) Zero-shot (ZS) -- no task-specific data. 
The approaches for ZS contain a sequence-to-sequence unified approach that is trained on multiple tasks \citep{lu2022unified, Zhu_2022_CVPR, gupta2022towards}. However, as the list of tasks is fixed, so any new task requires changes to the model and additional training.

{\em Socratic models}, an approach for few-shot and ZS learning, composes pre-trained models by directly using language as the intermediate representation by which the modules exchange information with each other\citep{zeng2022socratic}. Thus, this approach heavily relies on a large language model (LLM) and requires prompt engineering which does not have a proper methodology. 
Relying on LLMs might be sub-optimal, particularly for multimodal tasks that do not involve language, e.g., music and vision tasks \citep{qiu2018image, aleixo2021music}.

In this paper, we propose a different approach to multimodal tasks that leverages the expertise of foundation models and shares knowledge through a common latent space without relying on 
language as a mediator. 
The importance of knowledge sharing between experts can be illustrated by the Apollo program, which required the collaboration of experts from diverse fields, such as physics, chemistry, and biology, to achieve the common goal of landing a man on the moon. By sharing their knowledge, these experts were able to overcome the challenges and undertake a task never done before.
Our premise that complex tasks, like the Apollo, require multiple experts, inspired our approach which relies on synergy and knowledge sharing between pre-trained transformer components through gradient updating of a combined loss at inference time. This allows our model to perform new tasks in a zero-shot setup without any further training or tuning steps. 
Unlike Socratic models, the proposed framework, which we named {\scshape Apollo}, is not limited to language models. It can be applied to a variety of transformer models of different modalities, such as audio and vision, moving beyond LLMs and not depending on prompts. Furthermore, {\scshape Apollo} enables decentralized command execution, allowing each model to contribute and benefit from the expertise of others.

We demonstrate our approach on two tasks. On the well-known task of stylized image captioning \citep{MemCap:DBLP:conf/aaai/ZhaoWZ20, Guo2019MSCapMI, capdec:nukrai2022text, mathews2016senticap, gan2017stylenet}, our ZS Apollo method gained an absolute improvement of up to 58\% in style accuracy and up to 2.3\%  in relevance text to the image, compared to the state-of-the-art semi-supervised models on the SentiCap \citep{mathews2016senticap}  and FlickrStyle10K \citep{gan2017stylenet} benchmarks. 
We further demonstrate this method on a novel task, audio-aware image captioning, in which an image and audio are given and the task is to generate text that describes the image within the context of the provided audio.

\subsection{The {\scshape Apollo} methods}
\label{sec:Apollo}

The cutting-edge models across diverse modality domains primarily rely on transformer-based architectures \citep{vaswani2017attention}. 
Our objective is to leverage the expertise of multiple pre-trained transformer models to generate output through shared impact between the models.
A Transformer model consists of two primary components: an encoder and a decoder. Each component comprises $L$ layers of encoders and decoders, and within these layers, multiple attention heads are present, each with query ($Q$), key ($K$), and value ($V$) functions. 
The attention mechanism enables the model to selectively focus on different parts of the input data. This focus is determined by the interactions between $Q$ and $K$, which produce attention scores and influence the distribution of $V$. Function $Q$ operates on the input token embedding, while $K$ and $V$ generate subsequent output tokens by considering past tokens. This implies that both the $K$ and the $V$ can influence the final prediction output, given $Q$. To exercise control over the model's output, we seek to influence the 'context cache', which contains both the key ($K$) and the value ($V$), thus guiding the model's predictions towards a desired direction.  We consider a probability vector for the output of a transformer model $T_j$:
\text{$P_{T_{j}}(\{x_i\}_{i=1}^n | \{m_i\}_{i=1}^M ; T_j(\cdot | C^l_{T_{j}}))
$}, where $P_{T_{j}}$ represents the probability of candidates $\{x_i\}_{i=1}^n$ conditioning on modalities $\{m_i\}_{i=1}^M$. The probability is parameterized by an expert transformer $T_{j}$ for which we select a subset of $K$ and $V$ from certain layers $l$ to define a context, $C^l_{T_j}$.

\paragraph{Two Experts}
We generalize the loss function used by \cite{tewel2021zero} to any two  
 transformer models where transformer $T_{1}$ shares knowledge with ${T_{2}}$. We get the following loss:

\begin{equation}
    \mathcal{L} \triangleq CE(P_{T_{2}}^{(t)}, P_{T_{1}}) + \lambda \cdot CE(P_{T_{2}}^{(t)}, P_{T_{2}}^{(0)})
\label{eq:two_experts}
\end{equation}

In order to guide the model's prediction, we minimize the loss in  \eqref{eq:two_experts} over the context ${C_T}_2$, which implements the following concept:
The first term in  \eqref{eq:two_experts} pulls the preference tokens of transformer $T_2$ towards the target token preferences of $T_{1}$ through $t$ gradient steps, potentially overriding the original knowledge of transformer $T_2$. To preserve the transformer's original knowledge, an additive regularization term constrains the transformer's deviation from its initial preference, $P_{T_2}^{(0)}$.
$\lambda$ is a hyper-parameter that balances the two loss terms.
The guidance method implemented by  \eqref{eq:two_experts} is denoted as \textit{Experts-Summation}.

\paragraph{Multiple Experts}
We consider a framework that contains M$\geq$2 expert-transformers $\{T_{j}\}_{j=1}^M$. The \textit{Experts-Summation} can be extended to the multi-expert case by simply summing multiple weighted terms in the loss function:
\textbf{$\mathcal{L} = CE(P_{T_{M}}^{(t)}, P_{T_{M}}^{(0)}) + \Sigma_{j=1}^{M-1} \lambda_j\cdot CE\left(P_{T_M}^{(t)}, P_{T_{j}}\right)$}.
This extension comes at the cost of tuning multiple hyper-parameters, making it challenging to find the balance between all experts' loss components.
Therefore, we propose a new guidance loss inspired by the attention concept, which offers a safer alternative -- \textit{Experts-Product}:
\begin{equation}
     \mathcal{L} = CE({P_{T_{M}}}^{(t)}, {P_{T_{M}}}^{(0)}) +\lambda \cdot CE\left({P_{T_M}}^{(t)}, \Pi_{j=1}^{M-1} P_{T_{j}}\right)
 \label{eq:loss_prod_expert_source}
\end{equation}
The target probability in the second term of  \eqref{eq:loss_prod_expert_source} is the element-wise multiplication of all experts' probabilities, denoted as $\{P_{T_{j}}\}_{j=1}^{M-1}$.
This operation merges the experts' preferences and directs the transformer $T_M$ toward a common region, while maintaining proximity to the initial suggestion boundaries, as guided by the first loss term.
It does not add hyper-parameters comparing to \textit{Experts-Summation} and yet it effectively enforces ${P_T}_M^{(t)}$ to agree with the experts common support.

\begin{figure*}[t]
\centering
\scalebox{0.8}{
\includegraphics[width=\textwidth]
{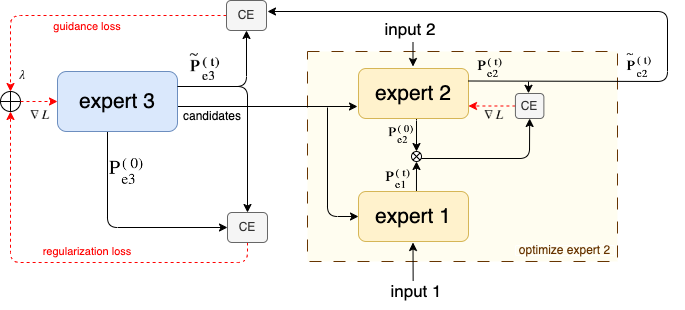}}
 \caption{An overview  of \textit{Decentralization of Guidance Efforts} approach.}
 \label{Decentralization_diagram}
 \end{figure*}

\paragraph{Decentralization of Guidance Efforts}
In the case of M expert-transformers, the straightforward way to apply all the experts' preferences to $P_T$ is by optimizing a flat objective function, as in \textit{Expert-Product}.
One challenge in accommodating all preferences simultaneously is the lack of effective communication among the guiding experts themselves.
Alternatively, we propose a hierarchical optimization process, 
in which one domain expert guides another, and the latter guides the top-level expert model. This allows experts to share their knowledge not only with the top-level expert model but also with each other.
In this process, a mediator expert is responsible for producing the final recommendation for the top-level expert model. This expert considers the perspective of the other experts and adapts to minimize potential conflicts in their guidelines.
To better understand this approach, 
we demonstrate it on a case of $M=3$ expert transformers as presented in Figure \ref{Decentralization_diagram}. In this example, expert 1 (${e_{1}}$) and 2 (${e_{2}}$) are domain-experts who guide a top-level expert - (${e_{3}}$) which plays a central role in the system.
$P_{e_{1}}, P_{e_{2}}, P_{e_{3}}$ denote the probabilities for the candidates $\{x_i\}_{i=1}^n$ over experts 1,2,3 respectively. 
The objective of aligning Expert 3 ($P_{e_{3}}$) with both Expert 1 ($P_{e_{1}}$) and Expert 2 ($P_{e_{2}}$) is achieved by solving the hierarchical optimization problems defined by the following equations:

\begin{equation}
                \Tilde{C}_{2}^{(t)} = \underset{C_2}{argmin}\{CE(P_{e_{2}}^{(t)}, P_{e_{2}}^{(0)} \cdot P_{e_{1}}^{(t)})\}
    \label{eq:optimize_e2}
\end{equation}

\begin{equation}
        \Tilde{C}_{3}^{(t)} = \underset{C_3}{argmin}\{CE(P_{e_{3}}^{(t)}, P_{e_{3}}^{(0)})+\lambda \cdot CE(P_{e_{3}}^{(t)}, \Tilde{P}_{e_{2}}^{(t)})\}
    \label{eq:optimize_e3}
\end{equation}
First, we optimize the probability $P_{e_{2}}$ over context cache $C_2$ (\eqref{eq:optimize_e2}). Second, we optimize the model probability $P_{e_{3}}$ at the top-hierarchy by adjusting $C_3$ (\eqref{eq:optimize_e3}). This approach decentralizes the guidance efforts among multiple models, enhancing the interaction between the experts.

\section{Stylized Image Caption Generation} \

\paragraph{Goal}
Our objective in this task is to generate captions that accurately describe the input image while incorporating the desired style. We aim to achieve this without training any model. Instead, our approach focuses on leveraging the expertise of diverse models and utilizing their capabilities to generate captions with the desired style.

\subsection{Method}

\paragraph{{\scshape Apollo-Cap}}
In order to generate captions for images with a specific style, we use multiple experts. We use the LLM GPT-2 \citep{GPT_2_Radford2019LanguageMA} to iteratively predict tokens. We use GPT-2 instead of its advanced versions, e.g., GPT-3, because GPT-2 is open-source, allowing us to modify its internal representations, such as its keys (Q) and values (V). We use an image-text alignment model -- CLIP \citep{CLIP:radford2021learning} to evaluate the relevance of each candidate token to the given image. Each candidate token is appended to the current partial sentence ($X_{t,i} = x_{i+1}^{(t)}, x_i,...,x_0$), and combined with the image as input to CLIP. The cosine similarity $S^{CLIP}$ between each candidate and the image is computed in the embedding space, and probabilities are generated by applying softmax with a smoothing temperature parameter $\tau$.
\begin{equation}
\begin{aligned}[b] 
  &  \bm S^{CLIP}_{x_{i+1}} = S^{CLIP}(\bm X_{t,i},I)\\
  &  \bm p_{x_{i+1}}^{CLIP} \triangleq softmax(\bm S^{CLIP}_{x_{i+1}};\tau)
        \label{eq:clip_prob}
\end{aligned}
\end{equation}
We consider the first layer output K and V as CLIP's \textit{context} for guidance purposes. Our last expert is a Style-Text Alignment
We employ a style classification model
and score each candidate based on its alignment with the desired style. We generate probabilities for all candidates by applying softmax with a smoothing temperature parameter. 
We use roBERTa \citep{liu2019roberta} for sentiment realization and DeepMoji \citep{felbo2017using} for applying romantic and humorous style.

\begin{equation}
\begin{aligned}[b] 
  &  \bm S^{STYLE}_{x_{i+1}} = S^{STYLE}(\bm X_{t,i}|STYLE)\\
  &  \bm p_{x_{i+1}}^{STYLE} 
        \triangleq softmax(\
        \bm S^{STYLE}_{x_{i+1}};\tau)
        \label{eq:style_prob}
\end{aligned}
\end{equation}

\begin{figure}[ht]
\raggedleft 
\noindent
\begin{minipage}{0.43\textwidth}
  \begin{algorithm}[H]
    \SetAlgoLined
    \DontPrintSemicolon
    \For{i=0...}{
        \For{t=0...T-1}{
            $\bm p_{x_{i+1}}^{(t)} \leftarrow GPT(\bm x_i,C_i^{(t)})$ \label{alg:row:GPT's predicition}\\ 
            $\bm X_{t,i} \leftarrow \bm x_{i+1}^{(t)}, x_i,...,x_0$ \\
            $\bm \{\bm p_{x_{i+1}}^{<expert>}\} \leftarrow 
            \text{calc\_probability}(\bm X_{t,i},I,STYLE)$\label{alg:row:calc_probability} \\
                $\mathcal{L} \leftarrow \text{calc\_loss}(\bm p_{x_{i+1}}^{(t)},\{\bm p_{x_{i+1}}^{<expert>}\},\bm p_{x_{i+1}}^{(0)})$\;\label{alg:row:calc_loss}
            
            $C_i^{(t+1)} \leftarrow{} C_i^{(t)} + \alpha\frac{\nabla_{C_i} \mathcal{L}}{\lVert \nabla_{C_i}\mathcal{L}\rVert ^2}$
            \label{alg:row:gradient_step}\\
        }
        $x_{i+1}  \leftarrow \underset{\bm x}{argmax} GPT(x_i,C_i^{(T)})$\\
        \If {$x_{i+1} =  EndToken$}{
            break
            }
    }
    \caption{Optimizing GPT-2 towards an image and style}
    \label{alg-APOLLO-Cap-general}
  \end{algorithm}
\end{minipage}
\hspace{4em}
\begin{minipage}{0.43\textwidth}
  \begin{algorithm}[H]
     \SetAlgoLined
    \DontPrintSemicolon
    Initialize $C^{(0)}$ to CLIP's default context \;
    \For{j=0...J-1}{
        $\bm P_{CLIP}^{(j)} = CLIP(\bm x, I | C^{(j)})$
        \\$\begin{aligned}
            \mathcal{L} \leftarrow CE\left(\bm P_{CLIP}^{(j)} , \bm P_{target}\right) \\
        \end{aligned} $
        $C^{(j+1)} \leftarrow{} C^{(j)} + \alpha\frac{\nabla_{C} \mathcal{L}}{\lVert \nabla_{C}\mathcal{L}\rVert ^2}$
        }
        \Return $CLIP(\bm x,I|C^{(J)})$
\caption{Optimizing CLIP image embedding towards the desired style}
\label{alg:optimize_CLIP_K_V}
  \end{algorithm}
\end{minipage}
\end{figure}

\subsection{Gradient Updates for Model Guiding} 
By combining the image-oriented and style-oriented probabilities, we can manipulate GPT-2 through its context vector to generate an image caption with the desired style.

Let $I$ be the input image, $x_{i+1}$ the next candidate token, $C_i$ the GPT-2's context vector, and $p_{x_{i+1}} = GPT(x_i, C_i)$  the probability predicted by GPT-2 for $x_{i+1}$. The goal is to iteratively optimize the context $C_i$ in order to improve the description of the image with the desired style. The optimization steps are outlined in Algorithm \ref{alg-APOLLO-Cap-general}.
For each generated token, a total of $T$ optimization steps are performed as follows:
An alternative probabilities of the next token are calculated according to a set of experts (row \ref{alg:row:calc_probability}).
Then, a loss function is computed incorporating the experts prediction (row \ref{alg:row:calc_loss}). As suggested by ZeroCap \citep{tewel2021zero}, a regularization term is added to keep the optimized probability close to the original probability generated by GPT-2 in the initial step.
Minimizing this loss over the context vector results in an image-style-aware probability.
The context vector is updated by applying a single gradient step (row \ref{alg:row:gradient_step}).
This optimization loop is repeated for each generated token until the captioning process is complete.
The outer loop is executed with 5 beams, and the inner loop is applied to the top K=512 tokens.

Next, we provide a detailed implementation for each guidance approach described in Section\ref{sec:Apollo}.

\paragraph{{\scshape Apollo-Cap}: Sum of Experts}
\label{sssec:Apollo-Cap:Sum of Experts}
After the generative transformer calculates its probability for the next token, each expert calculates its alternative probability.
To align image and text, we calculate the CLIP probability $\bm p_{x_{i+1}}^{CLIP}$ for the top 512 candidates (see \eqref{eq:clip_prob}) to determine the best probability vector for image-text correspondence.
In addition, style-aware probability $\bm p_{x_{i+1}}^{STYLE}$ is computed based on the style model's scores to encourage a certain style (see \eqref{eq:style_prob}).          
The guidance loss $\mathcal{L}$ is computed as a weighted sum of the cross-entropy between the augmented probabilities and the baseline GPT-2 probability:

\begin{equation}
    \begin{aligned}
        \mathcal{L} = 
         \lambda_{LM} CE\left(\bm p_{x_{i+1}}^{(t)},\bm p_{x_{i+1}}^{(0)}\right) 
         + \lambda_{CL} CE\left(\bm p_{x_{i+1}}^{(t)},\bm p_{x_{i+1}}^{CLIP}\right) 
        + \lambda_{SL} CE\left(\bm p_{x_{i+1}}^{(t)},\bm p_{x_{i+1}}^{STYLE}\right)  
    \end{aligned} 
\label{eq:loss-sum-of-experts}
\end{equation}

\paragraph{{\scshape Apollo-Cap}: Product of Experts}
\label{sssec:appolo-cap-Product of Experts}
Similarly to sum of experts, CLIP probability $\bm p_{x_{i+1}}^{CLIP}$ and style probability $\bm p_{x_{i+1}}^{STYLE}$ are computed according to \eqref{eq:clip_prob} and \eqref{eq:style_prob} respectively.
The guided loss $\mathcal{L}$ is composed of two terms: 
(1) the cross entropy between the product of CLIP and STYLE probabilities with the current GPT suggestion, and (2) a regularization term:
\begin{equation}
    \begin{aligned}
        \mathcal{L} = \
        & \lambda_{LM} \underbrace{ CE\left(\bm p_{x_{i+1}}^{(t)},\bm p_{x_{i+1}}^{(0)}\right)}_{regularization} 
         + \lambda_{CL} \underbrace{CE\left(\bm p_{x_{i+1}}^{(t)},\bm p_{x_{i+1}}^{CLIP} \cdot \bm p_{x_{i+1}}^{STYLE}\right)}_{experts} \\
    \end{aligned}
\label{eq:loss-prod-of-experts}
\end{equation}

\paragraph{{\scshape Apollo-Cap}: Decentralization}
\label{sssec:appolo-cap-decentralization}
We suggest optimizing CLIP's image embedding such that the resulting text-image matching will be more style-oriented.
Since CLIP is a transformer encoder, we apply the decentralization concept described in Section \ref{sec:Apollo} as follows:
Let $\bm x$ be candidate captions for image $I$.
We denote CLIP's first layer $K,V$ outputs by $C$ as context vector for optimization. Let $\bm P_{STYLE}$ be the probability vector produced by the style expert model given $\bm x$, and $\bm P_{CLIP}^{(0)} = CLIP(\bm x, I | C^{(0)})$ be CLIP's initial probability prediction for $\bm x$ given $I$ conditioning on the initial context $C^{(0)}$. We compute the target probability as the product of the style expert probability and CLIP's initial probability: $\bm P_{target} =  \bm P^{(0)}_{CLIP} \cdot \bm P_{STYLE}$.
We apply $J$ gradient steps to optimize CLIP's image embedding.
As a result, the optimized CLIP produces higher probabilities for captions that fit the image content from the specific style perspective. This approach is presented in Algorithm \ref{alg:optimize_CLIP_K_V}. 
 We denote the output probability as $\bm p_{x_{i+1}}^{CLIP-STYLE}$, and then incorporate it into the loss function  presented in 
  \eqref{eq:loss-prod-of-experts},
 resulting in the guidance loss $\mathcal{L}$:

\begin{equation}
    \begin{aligned}
\mathcal{L} =\
    &\lambda_{LM} CE\left(\bm p_{x_{i+1}}^{(t)},\bm p_{x_{i+1}}^{(0)}\right) 
     + \lambda_{CL} CE\left(\bm p_{x_{i+1}}^{(t)},\bm p_{x_{i+1}}^{CLIP-STYLE}\cdot \bm p_{x_{i+1}}^{STYLE}\right) 
    \end{aligned}
\label{eq:loss-decentralization}
\end{equation}

\subsection{ Experimental Setup}
\paragraph{Data} 
We evaluate our approach on the two benchmarks, SentiCap \citep{mathews2016senticap}  for positive and negative styling and FlickrStyle10K \citep{gan2017stylenet} for humor and romantic.

\paragraph{Evaluation Metrics}
\label{ssec:evaluation-metrics}
To evaluate the results, we examined the following attributes of the captions: (1) \emph{fluency}, i.e., the coherency and naturalness of the generated text; (2) \emph{Text-Image correspondence (TIC)}, and (3) \emph{style accuracy}.  We evaluate \emph{fluency}  using the perplexity function of GPT-2, which measures the model's ability to predict the next word in a sequence. Lower perplexity values indicate better fluency of the generated captions. The perplexity scores were clipped to the maximal value of 1500 and then normalized by $1-\dfrac{perplexity}{1500}$, formalizing a fluency score (the higher the better). In order to quantify the alignment between an image and its caption (\emph{TIC}), we used CLIPScore \citep{hessel2022clipscore} - the cosine similarity between the CLIP embedding of the image and the caption. We measured \emph{style accuracy}
using large pre-trained models -- roBERTa \citep{roberta:2022} and DeepMoji \citep{felbo2017using}. roBERTa is a sentiment classification model that generates probability for either positive or negative.
In order to evaluate the emotional styles of Flickrstyle10k -- humorous and romantic, we employed DeepMoji model. Given a text input, DeepMoji generates a 64-dimensional probability vector for various emotions which are aggregated to represent humorous and romantic styles (see Appendix \ref{sec:appendix-deepmoji}).

\paragraph{Models}
We demonstrated the three zero-shot methods described in Sections \ref{sssec:Apollo-Cap:Sum of Experts} by plugging-in the loss functions in \eqref{eq:loss-sum-of-experts},\eqref{eq:loss-prod-of-experts}, \eqref{eq:loss-decentralization} into ZeroCap as drop-and-replace of its original loss.
Specifically, we employed the techniques  \textit{Experts-Summation} which will be referred to as {\scshape Apollo-Cap}, \textit{Expert-Product} ({\scshape Apollo-Cap-P}) and combination of \textit{Decentralization of Guidance Efforts} with \textit{Expert-Product} ({\scshape Apollo-Cap-PD}).

\begin{table*}[t]
\centering
\scalebox{0.99}{
\scriptsize\begin{tabular}{ccccccccc}
  
                                          & \multicolumn{3}{c}{positive}                          &               & \multicolumn{4}{c}{negative}           \\
                                          \cmidrule(rl){2-5} \cmidrule(rl){6-9}
                                          
Model                                    & TIC      & style accuracy & fluency       & Vocab         & TIC      & style accuracy & fluency & Vocab \\ \hline
CapDec                                    & 0.294          & 0.53                & \textbf{0.97} & 717           & 0.292           & 0.3                 & \textbf{0.97}    & 706   \\ \hline

ZeroCap+{\scshape PM}
           & 0.327          & 0.79                & 0.93          & 2715          & 0.31             & 0.79                 & 0.94    & 2937  \\
ZeroCap+{\scshape IM}              & \textbf{0.328}          & 0.24                & 0.84          & \textbf{2917}          & \textbf{0.33}          & 0.13                 & 0.83    & 2884  \\
ZeroCap+{\scshape IPM}
 & 0.327          & 0.86                & 0.93          & 2736          & 0.312          & 0.8                 & 0.94    & \textbf{3025}  \\ \hline
{\scshape Apollo-Cap}                                & 0.268          & 0.91                & 0.9           & 1978          & 0.267          & 0.76                 & 0.86    & 2417 \\
{\scshape Apollo-Cap-P}                & 0.283          & \textbf{0.97}       & 0.84         & 1658          & 0.291          & \textbf{0.88}                 & 0.85    & 2302  \\
{\scshape Apollo-Cap-PD} & 0.317 & 0.94                & 0.8           & 2200 & 0.296 & 0.81        & 0.85     & 2544  \\
\hline\\

                                           \multicolumn{1}{c}{}    
                                          & \multicolumn{4}{c}{humorous}                       & \multicolumn{4}{c}{romantic}                             \\ 
                                          \cmidrule(rl){2-5} \cmidrule(rl){6-9}
Model                                    & TIC      & style accuracy & fluency       & Vocab         & TIC      & style accuracy & fluency & Vocab \\ \hline
CapDec                                    & 0.285          & 0.05                & \textbf{0.98} & 885           & 0.285           & 0.12                 & \textbf{0.98}    & 822   \\ \hline
ZeroCap+{\scshape PM}
           & 0.325          & 0.06                & 0.88          & 2875          & 0.321             & 0.09                 & 0.87    & \textbf{2983}  \\
ZeroCap+{\scshape IM}              & \textbf{0.326}          & 0.05                & 0.81          & 2818          & \textbf{0.325}          & 0.07                 & 0.8    & 2855  \\
ZeroCap+{\scshape PIM}
 & 0.325          & 0.07                & 0.93          & 2531          & 0.317          & 0.14                 & 0.92    & 2728  \\ \hline
{\scshape Apollo-Cap}                                & 0.269          & 0.06                & 0.91           & \textbf{3001}          & 0.268          & 0.13                 & 0.84    & 2712 
\\{\scshape Apollo-Cap-P}                & 0.286          & \textbf{0.23}       & 0.90         & 2444          & 0.262          & 0.32                 & 0.88    & 2621  \\
{\scshape Apollo-Cap-PD} & 0.298 & 0.2                & 0.85           & 2774 & 0.28 & \textbf{0.37}        & 0.81     & 2527 
\\\hline
*TIC- text-image correspondence
\end{tabular}
}
\caption{Averaged Scores for CapDec,  ZeroCap Manipulations and Apollo-Cap Approaches}
\small{ZeroCap best fits the image content, but fails to generate style with only input manipulations. CapDec as a semi supervised model for image captioning shows fluent language but also weaker style realization. {\scshape Apollo-Cap-PD} outperforms the other approaches in the total image-text-style matching trade-off.}
\label{tab:results}
\end{table*}

\paragraph{Baselines}
We conducted a comparative analysis of our method with the current state-of-the-art technique for generating stylized image captions, namely CapDec \citep{capdec:nukrai2022text}. CapDec, a semi-supervised method, relies on training a decoder using stylized text to generate stylized captions. It achieves this by leveraging the shared embedding space of text and images in CLIP. Following CapDec's training protocol, we trained on SentiCap and Flickrstyle10k datasets until the validation set loss reached a plateau. 
Additionally, we compared our results to the ZeroCap model \citep{tewel2021zero}, which incorporates a style injection manipulation. We implemented three different manipulation techniques: (1) $\bold{ZeroCap+PM}$, in which the style is injected into the LLM via prompting. 
We used the following prompts: for a positive style --  \enquote{The beautiful image of a}; for a negative style -- \enquote{The disturbing image of a}; for a humorous style -- \enquote{The humorous image of a}; and for a romantic style -- \enquote{The romantic image of a}.
 (2) $\bold{ZeroCap+IM}$, in which the style is injected via images into the CLIP model. We perform arithmetic operations on the input image embedding by adding the CLIP embedding of an emoji that represents the desired style (e.g., a smiley emoji for positive sentiment), and subtracting a neutral emoji embedding to discard the attributes belonging to the emoji itself. Finally, we implemented (3) $\bold{ZeroCap+IPM}$, a combination of both aforementioned manipulation techniques. 
Although these methods are based on a zero-shot model, they require careful selection of prompts and images to achieve the desired style.

\begin{figure*}[th]
\scalebox{0.99}{
\includegraphics[width=\textwidth]{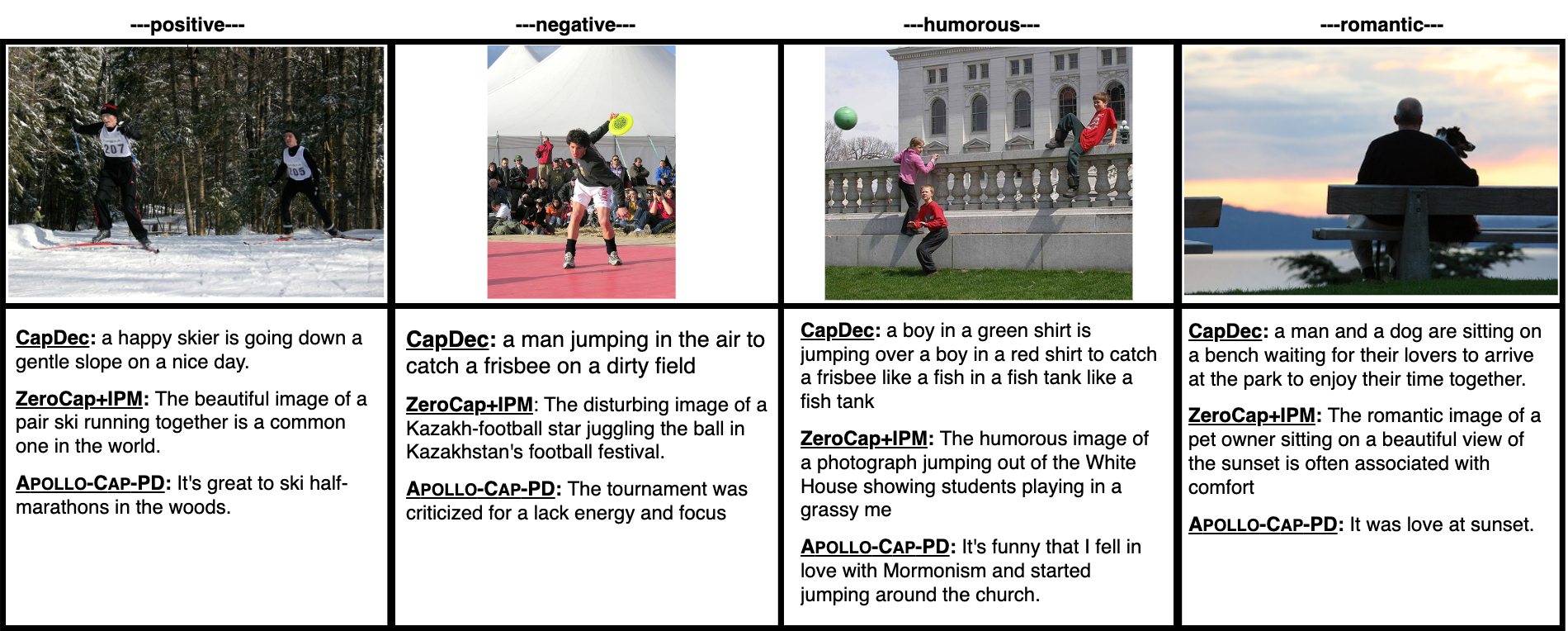}}
 \caption{Examples of our {\scshape Apollo-Cap-PD} compared to SOTA models.
 } 
\label{fig:instroctor}
\end{figure*}

\subsection{Results}

\paragraph{Quantitative analysis} 
Table \ref{tab:results} shows our results for SentiCap (top table) and Flickrstyle10k (bottom table).
The {\scshape Apollo-Cap}-based models outperformed all baselines in terms of style accuracy across all benchmarks.
Although the ZeroCap-based approaches gained the highest TIC scores, they were partially successful in generating the required style, and in the qualitative test hereafter they performed the worst compared to the other approaches.
The results also show that {\scshape Apollo-Cap-PD} surpassed the state-of-the-art model, CapDec, in style accuracy on all styles, and in TIC on all styles except for the romantic style, while only slightly reducing the fluency score. It is important to note that this minor impact on fluency is acceptable, as a fluency score of 0.8 already indicates a good fluency level. 
Upon observing the results based on \textsc{Apollo-Cap}, we can see that \textsc{Apollo-Cap-P} and \textsc{Apollo-Cap-PD} achieve significantly higher results in TIC and style accuracy, than \textsc{Apollo-Cap}.  \textsc{Apollo-Cap-PD}  outperforms  \textsc{Apollo-Cap-P} on TIC across all styles,  but it is unclear which method \textsc{Apollo-Cap-P} or \textsc{Apollo-Cap-PD} performs better on the style accuracy. 
Additionally, the fluency scores for all of these approaches are sufficient, exceeding 0.8.
The ZS methods based on \textsc{Apollo-Cap} and ZeroCap exhibit larger vocabularies than the CapDec, which was trained on the task-specific dataset.

\paragraph{Qualitative Analysis}  
In Figure \ref{fig:instroctor} we present a comprehensive comparison of several approaches: {\scshape Apollo-Cap-PD} (our leading approach), CapDec, and ZeroCap+IPM. We show results for the styles: positive, negative, humorous, and romantic. 
When comparing \textsc{Apollo-Cap-PD} to CapDec, we observed that the former exhibits broader world knowledge in its captions, while the latter focuses mainly on technical details. For example, in the negative caption, \textsc{Apollo-Cap-PD} identified the scene as a tournament, whereas CapDec provided drier factual information (\enquote{a man jumping...}). Moreover, CapDec used mainly common adjectives (e.g., `dirty') to embed the style. In contrast, \textsc{Apollo-Cap-PD} provided creative descriptions, such as  `criticized for a lack of energy', which contextualize the style within the narrative. This difference may be explained by CapDec mimicking the limited style displayed in training data, while \textsc{Apollo-Cap-PD} leverages a general LLM that naturally implements styles in a storyline.

\begin{figure}[t]
    \centering

    \begin{minipage}[t]{0.45\linewidth}
        \centering
        \includegraphics[width=\linewidth]{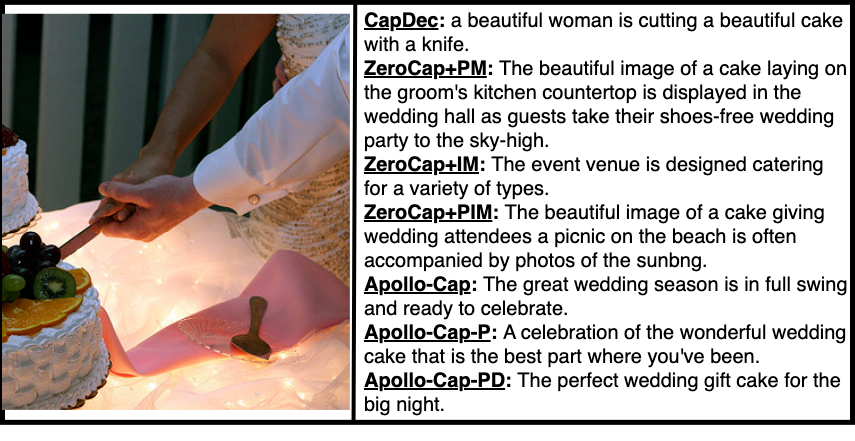}
        \caption{A positive-style caption example.}
        \label{fig:ex-all_approaches}
    \end{minipage}\hfill   
    \begin{minipage}[t]{0.45\linewidth}
        \centering
        \includegraphics[width=\linewidth]{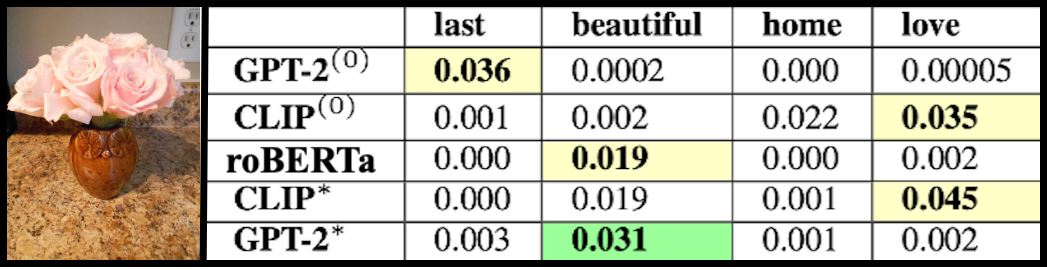}
        \caption{Models' token probabilities  
        }
        \label{fig:apollo-explained}
    \end{minipage}
\end{figure}

\paragraph{Ablation Study} 
Figure \ref{fig:ex-all_approaches} provides a comprehensive comparison of all approaches, illustrating a positive image caption.
CapDec properly described the fact that a woman is cutting a cake and also added positive adjectives, yet it lacked real-world knowledge. In this case, it missed the celebration context.
While ZeroCap's approach captured some relevant details 
they also  
exhibited instances of hallucination, as seen in examples like `shoes-free wedding' and `picnic beach'. 
{\scshape Apollo-Cap} expressed the celebration of the wedding, and yet it dropped an important part of the content - the cake.
In comparison, {\scshape Apollo-Cap-P} included the important details - the celebration, the wedding, the cake, and relevant style adjectives; however, the fluency is degraded.
Finally, {\scshape Apollo-Cap-PD}  met all the criteria - relevance (`wedding', `cake' and `night'), style (`perfect'), and fluency that is reflected in the proper integration of the content words.

\paragraph{Model Ensemble Analysis}
We provide a detailed explanation of the optimization process of {\scshape Apollo-Cap-PD} for an input image of a vase with roses and a desired positive style. In this section, the notations GPT$^{(0)}$ and GPT$^{(*)}$, CLIP$^{(0)}$ and CLIP$^{(*)}$  refer to algorithms \ref{alg-APOLLO-Cap-general} and \ref{alg:optimize_CLIP_K_V}, respectively. The superscript zero denotes the expert model with its original context vector and the asterisk denotes the expert model with its optimized context vector. 
After generating the first token, `The', Figure \ref{fig:apollo-explained}, displays the probability of the top-1 candidate tokens according to each expert. GPT-2$^{(0)}$ assigns the highest probability to `last', while CLIP$^{(0)}$ identified `love' as the next most likely token, possibly due to the common association of vases with roses in a romantic context. In contrast, roBERTa preferred `beautiful' as the next token, possibly indicating its frequent use as a positive adjective to start a sentence.
After one optimization step, CLIP$^{(*)}$ maintained its original preferences but also increased the probability of `beautiful', aligning with both the desired style and the image. Finally, with the combined loss propagation of GPT-2, CLIP and roBERTa expert models, GPT-2$^{(*)}$ selected `beautiful' as the token, which appears to be the most probable choice in this context. This suggests that {\scshape Apollo-Cap-PD} considers various aspects, including style, image relevance, and text fluency.

\begin{figure}[t]
\scalebox{0.99}{
\includegraphics[width=\textwidth]{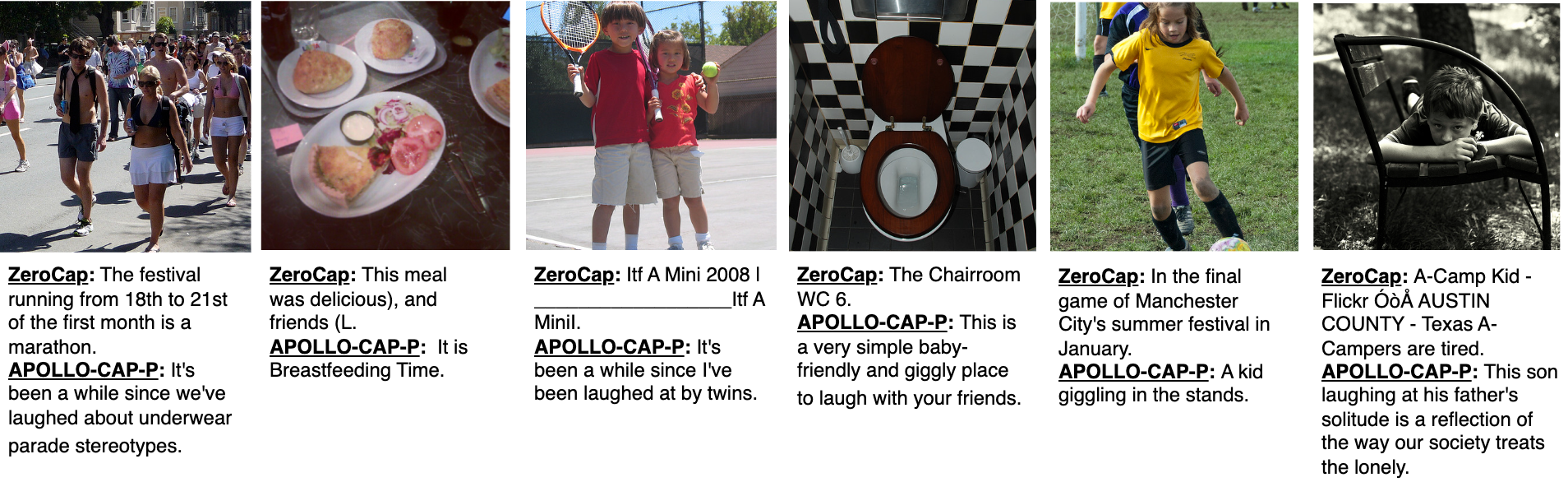}}
 \caption{{\scshape Apollo-Cap-P} caption examples for images and audio clips featuring children's laughter.
 } 
\label{fig:apollo-cap-clap}
\end{figure}

\section{Audio-Aware Image Captioning}
\paragraph{Goal} We demonstrate our approach's ability to generalize to other modalities by introducing a novel task -- Audio-Aware Image Captioning, that integrates audio into image captions. The input of the task is both an image and an audio clip, and the task is to generate text that describes the image within the context of the provided audio.

\paragraph{Data} The test set contains 50 randomly sampled images from the Senticap \citep{mathews2016senticap} test set, and the validation set contains five images from the Senticap validation set. For all images, we included an audio clip of children’s laughter that we collected from \url{https://freesound.org}\footnote{The audio is available at \url{https://github.com/danielabd/Apollo-Cap}}.

\paragraph{Model} We adapted the stylized image captioning system of {\scshape Apollo-Cap-P} (Section \ref{sssec:appolo-cap-Product of Experts}) by replacing the style component with an audio counterpart -- CLAP \citep{laionclap2023}, which assess the correspondence between text and audio. We projected the audio clip and the candidate captions onto the CLAP embedding space and calculated the cosine similarity between each candidate and the audio embedding vectors. We then replaced the style probability in  \eqref{eq:style_prob} with the audio probability and applied the rest of the algorithm without further changes.

\paragraph{Qualitative Analysis}
Figure \ref{fig:apollo-cap-clap} shows examples of captions generated by {\scshape Apollo-Cap-P} for images from the test set in the presence of audio featuring kids' laughter. 
{\scshape Apollo-Cap-P}  managed in all images to add the context of the audio -- laughing. ZeroCap, which does not process audio, does not reflect this context. In the first left image {\scshape Apollo-Cap-P}  reasons that an image with many people walking is a parade and because they are with very little clothing he makes it into something funny (connected with the audio) -- an `underwear parade'. The second image humorously connects the image of a mother eating with the sound of a baby laughing, suggesting that the baby is eating as well, since the mother is eating -- \enquote{It is Breastfeeding Time}. Even when the image seems gloomy as in the rightmost image, the model manages to generate a caption that connects laughter with the negative sentiment of the image. 
These results demonstrate our method's ability to process audio and still show scene-level understanding.

\begin{table}
\centering
    \begin{tabular}{lccr}
         \textbf{Model} & \textbf{TAC} & \textbf{TIC} & \textbf{Fluency}\\
         \midrule
         {\scshape Apollo-Cap-P} & 0.53&0.28 & 0.91 \\
         ZeroCap  & 0.08&0.32 & 0.81 \\
         \bottomrule\\
         
    \end{tabular}
    \caption{Averaged Scores for ZeroCap and {\scshape APOLLO-CAP-P} with laughter audio content on 50 images from the Senticap test set. 
    TIC - text-image correspondence, TAC - text-audio correspondence}
\label{tab:audio_results}
\end{table}

\paragraph{Quantitative Analysis} Table \ref{tab:audio_results} shows our {\scshape Apollo-Cap-P} approach manages to get the text-audio correspondence (TAC), while maintaining the high fluency and text-image correspondence (TIC).

\section{Related Work}

In recent years, there has been a shift in modeling towards transformer-based methods \citep{vaswani2017attention} which learn context and process sequential data through their attention mechanism.
The next revolutions in machine learning came with the rise of \textit{foundation models}, which are transformer-based models that have been injected with prior knowledge through pre-training on large datasets \citep{devlin2018bert, lan2019albert, yang2019xlnet, zan2022cert, kim2021vilt, zaheer2020big, baevski2020wav2vec}. These models have been shown to perform on various tasks, domains, and even multimodality \citep{liu2023prismer,NEURIPS2022_00d1f03b}
Their distinct capabilities mainly depend on their training data, for example, models that are trained on pairs of images and texts demonstrate capabilities in vision and language tasks \citep{tan2019lxmert, lu2019vilbert, CLIP:radford2021learning}.

\textit{Foundation} models' capabilities to perform in zero-shot have been utilized in \textit{Socratic} approach which combines foundation models with frozen LLMs and bridges the gap through language, via prompting  \citep{zeng2022socratic, tiong2022plug, wang2022language, NEURIPS2021_01b7575c, huang2023visual, xie2022visual}. 
In contrast to Socratic models, a different approach, that does not rely on prompting, guides the LLMs by tuning their prior knowledge in their attention mechanism with visual cues  \citep{tewel2021zero}. In our work, we
present a generic approach for guiding multiple transformer models through gradient updates, which can be employed across different modalities. 

\section{Conclusions}
We propose a modular framework that leverages the expertise of large pre-trained models and jointly solves complex tasks in a zero-shot setting without relying on prompting. Our approach enables decentralized control, allowing models to exchange expertise. 
We demonstrated our approach on two tasks. Our method achieves state-of-the-art results on two benchmarks for stylized image captioning. 
To demonstrate the method's capabilities, we tested its ability to work on audio, by introducing the novel task of audio-aware image captioning, in which an image and audio are given and the task is to generate text that describes
the image within the context of the provided audio.
\bibliography{iclr2024_conference}

\begin{thebibliography}{47}
\providecommand{\natexlab}[1]{#1}
\providecommand{\url}[1]{\texttt{#1}}
\expandafter\ifx\csname urlstyle\endcsname\relax
  \providecommand{\doi}[1]{doi: #1}\else
  \providecommand{\doi}{doi: \begingroup \urlstyle{rm}\Url}\fi

\bibitem[Aleixo et~al.(2021)Aleixo, Pinto, and Correia]{aleixo2021music}
Lu{\'\i}s Aleixo, H~Sofia Pinto, and Nuno Correia.
\newblock From music to image a computational creativity approach.
\newblock In \emph{Artificial Intelligence in Music, Sound, Art and Design:
  10th International Conference, EvoMUSART 2021, Held as Part of EvoStar 2021,
  Virtual Event, April 7--9, 2021, Proceedings 10}, pp.\  379--395. Springer,
  2021.

\bibitem[Antol et~al.(2015)Antol, Agrawal, Lu, Mitchell, Batra, Zitnick, and
  Parikh]{antol2015vqa}
Stanislaw Antol, Aishwarya Agrawal, Jiasen Lu, Margaret Mitchell, Dhruv Batra,
  C~Lawrence Zitnick, and Devi Parikh.
\newblock Vqa: Visual question answering.
\newblock In \emph{Proceedings of the IEEE international conference on computer
  vision}, pp.\  2425--2433, 2015.

\bibitem[Baevski et~al.(2020)Baevski, Zhou, Mohamed, and
  Auli]{baevski2020wav2vec}
Alexei Baevski, Yuhao Zhou, Abdelrahman Mohamed, and Michael Auli.
\newblock wav2vec 2.0: A framework for self-supervised learning of speech
  representations.
\newblock \emph{Advances in neural information processing systems},
  33:\penalty0 12449--12460, 2020.

\bibitem[Brown et~al.(2020)Brown, Mann, Ryder, Subbiah, Kaplan, Dhariwal,
  Neelakantan, Shyam, Sastry, Askell, et~al.]{brown2020language}
Tom Brown, Benjamin Mann, Nick Ryder, Melanie Subbiah, Jared~D Kaplan, Prafulla
  Dhariwal, Arvind Neelakantan, Pranav Shyam, Girish Sastry, Amanda Askell,
  et~al.
\newblock Language models are few-shot learners.
\newblock \emph{Advances in neural information processing systems},
  33:\penalty0 1877--1901, 2020.

\bibitem[Chen et~al.(2019)Chen, Suhr, Misra, Snavely, and
  Artzi]{chen2019touchdown}
Howard Chen, Alane Suhr, Dipendra Misra, Noah Snavely, and Yoav Artzi.
\newblock Touchdown: Natural language navigation and spatial reasoning in
  visual street environments.
\newblock In \emph{Proceedings of the IEEE/CVF Conference on Computer Vision
  and Pattern Recognition (CVPR)}, June 2019.

\bibitem[Devlin et~al.(2018)Devlin, Chang, Lee, and Toutanova]{devlin2018bert}
Jacob Devlin, Ming-Wei Chang, Kenton Lee, and Kristina Toutanova.
\newblock Bert: Pre-training of deep bidirectional transformers for language
  understanding.
\newblock \emph{arXiv preprint arXiv:1810.04805}, 2018.

\bibitem[Felbo et~al.(2017)Felbo, Mislove, S{\o}gaard, Rahwan, and
  Lehmann]{felbo2017using}
Bjarke Felbo, Alan Mislove, Anders S{\o}gaard, Iyad Rahwan, and Sune Lehmann.
\newblock Using millions of emoji occurrences to learn any-domain
  representations for detecting sentiment, emotion and sarcasm.
\newblock \emph{arXiv preprint arXiv:1708.00524}, 2017.

\bibitem[Gan et~al.(2017)Gan, Gan, He, Gao, and Deng]{gan2017stylenet}
Chuang Gan, Zhe Gan, Xiaodong He, Jianfeng Gao, and Li~Deng.
\newblock Stylenet: Generating attractive visual captions with styles.
\newblock In \emph{Proceedings of the IEEE conference on computer vision and
  pattern recognition}, pp.\  3137--3146, 2017.

\bibitem[Guo et~al.(2019)Guo, Liu, Yao, Li, and Lu]{Guo2019MSCapMI}
Longteng Guo, Jing Liu, Peng Yao, Jiangwei Li, and Hanqing Lu.
\newblock Mscap: Multi-style image captioning with unpaired stylized text.
\newblock \emph{2019 IEEE/CVF Conference on Computer Vision and Pattern
  Recognition (CVPR)}, pp.\  4199--4208, 2019.

\bibitem[Gupta et~al.(2022)Gupta, Kamath, Kembhavi, and
  Hoiem]{gupta2022towards}
Tanmay Gupta, Amita Kamath, Aniruddha Kembhavi, and Derek Hoiem.
\newblock Towards general purpose vision systems: An end-to-end task-agnostic
  vision-language architecture.
\newblock In \emph{Proceedings of the IEEE/CVF Conference on Computer Vision
  and Pattern Recognition}, pp.\  16399--16409, 2022.

\bibitem[Hartmann et~al.(2022)Hartmann, Heitmann, Siebert, and
  Schamp]{roberta:2022}
Jochen Hartmann, Mark Heitmann, Christian Siebert, and Christina Schamp.
\newblock More than a feeling: Accuracy and application of sentiment analysis.
\newblock \emph{International Journal of Research in Marketing}, 40, 06 2022.
\newblock \doi{10.1016/j.ijresmar.2022.05.005}.

\bibitem[Hessel et~al.(2022)Hessel, Holtzman, Forbes, Bras, and
  Choi]{hessel2022clipscore}
Jack Hessel, Ari Holtzman, Maxwell Forbes, Ronan~Le Bras, and Yejin Choi.
\newblock Clipscore: A reference-free evaluation metric for image captioning,
  2022.

\bibitem[Huang et~al.(2023)Huang, Mees, Zeng, and Burgard]{huang2023visual}
Chenguang Huang, Oier Mees, Andy Zeng, and Wolfram Burgard.
\newblock Visual language maps for robot navigation.
\newblock In \emph{2023 IEEE International Conference on Robotics and
  Automation (ICRA)}, pp.\  10608--10615. IEEE, 2023.

\bibitem[Ji et~al.(2022)Ji, Kojima, Rush, Suhr, Vong, Hawkins, and
  Artzi]{ji2022abstract}
Anya Ji, Noriyuki Kojima, Noah Rush, Alane Suhr, Wai~Keen Vong, Robert~D
  Hawkins, and Yoav Artzi.
\newblock Abstract visual reasoning with tangram shapes.
\newblock \emph{arXiv preprint arXiv:2211.16492}, 2022.

\bibitem[Kim et~al.(2021)Kim, Son, and Kim]{kim2021vilt}
Wonjae Kim, Bokyung Son, and Ildoo Kim.
\newblock Vilt: Vision-and-language transformer without convolution or region
  supervision.
\newblock In \emph{International Conference on Machine Learning}, pp.\
  5583--5594. PMLR, 2021.

\bibitem[Ku et~al.(2020)Ku, Anderson, Patel, Ie, and Baldridge]{ku2020room}
Alexander Ku, Peter Anderson, Roma Patel, Eugene Ie, and Jason Baldridge.
\newblock {Room-Across-Room}: Multilingual vision-and-language navigation with
  dense spatiotemporal grounding.
\newblock In \emph{Conference on Empirical Methods for Natural Language
  Processing (EMNLP)}, 2020.

\bibitem[Lan et~al.(2019)Lan, Chen, Goodman, Gimpel, Sharma, and
  Soricut]{lan2019albert}
Zhenzhong Lan, Mingda Chen, Sebastian Goodman, Kevin Gimpel, Piyush Sharma, and
  Radu Soricut.
\newblock Albert: A lite bert for self-supervised learning of language
  representations.
\newblock \emph{arXiv preprint arXiv:1909.11942}, 2019.

\bibitem[Lei et~al.(2018)Lei, Yu, Bansal, and Berg]{lei2018tvqa}
Jie Lei, Licheng Yu, Mohit Bansal, and Tamara~L Berg.
\newblock Tvqa: Localized, compositional video question answering.
\newblock \emph{arXiv preprint arXiv:1809.01696}, 2018.

\bibitem[Li et~al.(2022)Li, Niu, and Zhang]{Li_2022_CVPR}
Jiangtong Li, Li~Niu, and Liqing Zhang.
\newblock From representation to reasoning: Towards both evidence and
  commonsense reasoning for video question-answering.
\newblock In \emph{Proceedings of the IEEE/CVF Conference on Computer Vision
  and Pattern Recognition (CVPR)}, pp.\  21273--21282, June 2022.

\bibitem[Liu et~al.(2023)Liu, Fan, Johns, Yu, Xiao, and
  Anandkumar]{liu2023prismer}
Shikun Liu, Linxi Fan, Edward Johns, Zhiding Yu, Chaowei Xiao, and Anima
  Anandkumar.
\newblock Prismer: A vision-language model with an ensemble of experts.
\newblock \emph{arXiv preprint arXiv:2303.02506}, 2023.

\bibitem[Liu et~al.(2019)Liu, Ott, Goyal, Du, Joshi, Chen, Levy, Lewis,
  Zettlemoyer, and Stoyanov]{liu2019roberta}
Yinhan Liu, Myle Ott, Naman Goyal, Jingfei Du, Mandar Joshi, Danqi Chen, Omer
  Levy, Mike Lewis, Luke Zettlemoyer, and Veselin Stoyanov.
\newblock Roberta: A robustly optimized bert pretraining approach, 2019.

\bibitem[Lu et~al.(2019)Lu, Batra, Parikh, and Lee]{lu2019vilbert}
Jiasen Lu, Dhruv Batra, Devi Parikh, and Stefan Lee.
\newblock Vilbert: Pretraining task-agnostic visiolinguistic representations
  for vision-and-language tasks.
\newblock \emph{Advances in neural information processing systems}, 32, 2019.

\bibitem[Lu et~al.(2022)Lu, Clark, Zellers, Mottaghi, and
  Kembhavi]{lu2022unified}
Jiasen Lu, Christopher Clark, Rowan Zellers, Roozbeh Mottaghi, and Aniruddha
  Kembhavi.
\newblock Unified-io: A unified model for vision, language, and multi-modal
  tasks.
\newblock \emph{arXiv preprint arXiv:2206.08916}, 2022.

\bibitem[Mathews et~al.(2016)Mathews, Xie, and He]{mathews2016senticap}
Alexander Mathews, Lexing Xie, and Xuming He.
\newblock Senticap: Generating image descriptions with sentiments.
\newblock In \emph{Proceedings of the AAAI conference on artificial
  intelligence}, volume~30, 2016.

\bibitem[Nukrai et~al.(2022)Nukrai, Mokady, and
  Globerson]{capdec:nukrai2022text}
David Nukrai, Ron Mokady, and Amir Globerson.
\newblock Text-only training for image captioning using noise-injected clip.
\newblock \emph{arXiv preprint arXiv:2211.00575}, 2022.

\bibitem[Paz-Argaman et~al.(2020)Paz-Argaman, Atzmon, Chechik, and
  Tsarfaty]{paz2020zest}
Tzuf Paz-Argaman, Yuval Atzmon, Gal Chechik, and Reut Tsarfaty.
\newblock Zest: Zero-shot learning from text descriptions using textual
  similarity and visual summarization.
\newblock \emph{arXiv preprint arXiv:2010.03276}, 2020.

\bibitem[Qiu \& Kataoka(2018)Qiu and Kataoka]{qiu2018image}
Yue Qiu and Hirokatsu Kataoka.
\newblock Image generation associated with music data.
\newblock In \emph{Proceedings of the IEEE Conference on Computer Vision and
  Pattern Recognition Workshops}, pp.\  2510--2513, 2018.

\bibitem[Radford et~al.(2019)Radford, Wu, Child, Luan, Amodei, and
  Sutskever]{GPT_2_Radford2019LanguageMA}
Alec Radford, Jeff Wu, Rewon Child, David Luan, Dario Amodei, and Ilya
  Sutskever.
\newblock Language models are unsupervised multitask learners.
\newblock 2019.

\bibitem[Radford et~al.(2021)Radford, Kim, Hallacy, Ramesh, Goh, Agarwal,
  Sastry, Askell, Mishkin, Clark, et~al.]{CLIP:radford2021learning}
Alec Radford, Jong~Wook Kim, Chris Hallacy, Aditya Ramesh, Gabriel Goh,
  Sandhini Agarwal, Girish Sastry, Amanda Askell, Pamela Mishkin, Jack Clark,
  et~al.
\newblock Learning transferable visual models from natural language
  supervision.
\newblock In \emph{International conference on machine learning}, pp.\
  8748--8763. PMLR, 2021.

\bibitem[Rassin et~al.(2023)Rassin, Hirsch, Glickman, Ravfogel, Goldberg, and
  Chechik]{rassin2023linguistic}
Royi Rassin, Eran Hirsch, Daniel Glickman, Shauli Ravfogel, Yoav Goldberg, and
  Gal Chechik.
\newblock Linguistic binding in diffusion models: Enhancing attribute
  correspondence through attention map alignment.
\newblock \emph{arXiv preprint arXiv:2306.08877}, 2023.

\bibitem[Su et~al.(2022)Su, Lan, Liu, Liu, Yogatama, Wang, Kong, and
  Collier]{su2022language}
Yixuan Su, Tian Lan, Yahui Liu, Fangyu Liu, Dani Yogatama, Yan Wang, Lingpeng
  Kong, and Nigel Collier.
\newblock Language models can see: Plugging visual controls in text generation.
\newblock \emph{arXiv preprint arXiv:2205.02655}, 2022.

\bibitem[Tan \& Bansal(2019)Tan and Bansal]{tan2019lxmert}
Hao Tan and Mohit Bansal.
\newblock Lxmert: Learning cross-modality encoder representations from
  transformers.
\newblock \emph{arXiv preprint arXiv:1908.07490}, 2019.

\bibitem[Tewel et~al.(2021)Tewel, Shalev, Schwartz, and Wolf]{tewel2021zero}
Yoad Tewel, Yoav Shalev, Idan Schwartz, and Lior Wolf.
\newblock Zero-shot image-to-text generation for visual-semantic arithmetic.
\newblock \emph{arXiv preprint arXiv:2111.14447}, 2021.

\bibitem[Tiong et~al.(2022)Tiong, Li, Li, Savarese, and Hoi]{tiong2022plug}
Anthony Meng~Huat Tiong, Junnan Li, Boyang Li, Silvio Savarese, and Steven~CH
  Hoi.
\newblock Plug-and-play vqa: Zero-shot vqa by conjoining large pretrained
  models with zero training.
\newblock \emph{arXiv preprint arXiv:2210.08773}, 2022.

\bibitem[Tsimpoukelli et~al.(2021)Tsimpoukelli, Menick, Cabi, Eslami, Vinyals,
  and Hill]{NEURIPS2021_01b7575c}
Maria Tsimpoukelli, Jacob~L Menick, Serkan Cabi, S.~M.~Ali Eslami, Oriol
  Vinyals, and Felix Hill.
\newblock Multimodal few-shot learning with frozen language models.
\newblock In M.~Ranzato, A.~Beygelzimer, Y.~Dauphin, P.S. Liang, and J.~Wortman
  Vaughan (eds.), \emph{Advances in Neural Information Processing Systems},
  volume~34, pp.\  200--212. Curran Associates, Inc., 2021.
\newblock URL
  \url{https://proceedings.neurips.cc/paper_files/paper/2021/file/01b7575c38dac42f3cfb7d500438b875-Paper.pdf}.

\bibitem[Vaswani et~al.(2017)Vaswani, Shazeer, Parmar, Uszkoreit, Jones, Gomez,
  Kaiser, and Polosukhin]{vaswani2017attention}
Ashish Vaswani, Noam Shazeer, Niki Parmar, Jakob Uszkoreit, Llion Jones,
  Aidan~N Gomez, {\L}ukasz Kaiser, and Illia Polosukhin.
\newblock Attention is all you need.
\newblock \emph{Advances in neural information processing systems}, 30, 2017.

\bibitem[Wang et~al.(2020)Wang, Tran, and Feiszli]{Wang_2020_CVPR}
Weiyao Wang, Du~Tran, and Matt Feiszli.
\newblock What makes training multi-modal classification networks hard?
\newblock In \emph{Proceedings of the IEEE/CVF Conference on Computer Vision
  and Pattern Recognition (CVPR)}, June 2020.

\bibitem[Wang et~al.(2022)Wang, Li, Xu, Zhou, Lei, Lin, Wang, Yang, Zhu, Hoiem,
  et~al.]{wang2022language}
Zhenhailong Wang, Manling Li, Ruochen Xu, Luowei Zhou, Jie Lei, Xudong Lin,
  Shuohang Wang, Ziyi Yang, Chenguang Zhu, Derek Hoiem, et~al.
\newblock Language models with image descriptors are strong few-shot
  video-language learners.
\newblock \emph{Advances in Neural Information Processing Systems},
  35:\penalty0 8483--8497, 2022.

\bibitem[Wu* et~al.(2023)Wu*, Chen*, Zhang*, Hui*, Berg-Kirkpatrick, and
  Dubnov]{laionclap2023}
Yusong Wu*, Ke~Chen*, Tianyu Zhang*, Yuchen Hui*, Taylor Berg-Kirkpatrick, and
  Shlomo Dubnov.
\newblock Large-scale contrastive language-audio pretraining with feature
  fusion and keyword-to-caption augmentation.
\newblock In \emph{IEEE International Conference on Acoustics, Speech and
  Signal Processing, ICASSP}, 2023.

\bibitem[Xie et~al.(2022)Xie, Zhou, Dai, Yuan, Bach, Liu, and
  Zeng]{xie2022visual}
Yujia Xie, Luowei Zhou, Xiyang Dai, Lu~Yuan, Nguyen Bach, Ce~Liu, and Michael
  Zeng.
\newblock Visual clues: Bridging vision and language foundations for image
  paragraph captioning.
\newblock \emph{Advances in Neural Information Processing Systems},
  35:\penalty0 17287--17300, 2022.

\bibitem[Yang et~al.(2022)Yang, Miech, Sivic, Laptev, and
  Schmid]{NEURIPS2022_00d1f03b}
Antoine Yang, Antoine Miech, Josef Sivic, Ivan Laptev, and Cordelia Schmid.
\newblock Zero-shot video question answering via frozen bidirectional language
  models.
\newblock In S.~Koyejo, S.~Mohamed, A.~Agarwal, D.~Belgrave, K.~Cho, and A.~Oh
  (eds.), \emph{Advances in Neural Information Processing Systems}, volume~35,
  pp.\  124--141. Curran Associates, Inc., 2022.
\newblock URL
  \url{https://proceedings.neurips.cc/paper_files/paper/2022/file/00d1f03b87a401b1c7957e0cc785d0bc-Paper-Conference.pdf}.

\bibitem[Yang et~al.(2019)Yang, Dai, Yang, Carbonell, Salakhutdinov, and
  Le]{yang2019xlnet}
Zhilin Yang, Zihang Dai, Yiming Yang, Jaime Carbonell, Russ~R Salakhutdinov,
  and Quoc~V Le.
\newblock Xlnet: Generalized autoregressive pretraining for language
  understanding.
\newblock \emph{Advances in neural information processing systems}, 32, 2019.

\bibitem[Zaheer et~al.(2020)Zaheer, Guruganesh, Dubey, Ainslie, Alberti,
  Ontanon, Pham, Ravula, Wang, Yang, et~al.]{zaheer2020big}
Manzil Zaheer, Guru Guruganesh, Kumar~Avinava Dubey, Joshua Ainslie, Chris
  Alberti, Santiago Ontanon, Philip Pham, Anirudh Ravula, Qifan Wang, Li~Yang,
  et~al.
\newblock Big bird: Transformers for longer sequences.
\newblock \emph{Advances in neural information processing systems},
  33:\penalty0 17283--17297, 2020.

\bibitem[Zan et~al.(2022)Zan, Chen, Yang, Lin, Kim, Guan, Wang, Chen, and
  Lou]{zan2022cert}
Daoguang Zan, Bei Chen, Dejian Yang, Zeqi Lin, Minsu Kim, Bei Guan, Yongji
  Wang, Weizhu Chen, and Jian-Guang Lou.
\newblock Cert: Continual pre-training on sketches for library-oriented code
  generation.
\newblock \emph{arXiv preprint arXiv:2206.06888}, 2022.

\bibitem[Zeng et~al.(2022)Zeng, Wong, Welker, Choromanski, Tombari, Purohit,
  Ryoo, Sindhwani, Lee, Vanhoucke, et~al.]{zeng2022socratic}
Andy Zeng, Adrian Wong, Stefan Welker, Krzysztof Choromanski, Federico Tombari,
  Aveek Purohit, Michael Ryoo, Vikas Sindhwani, Johnny Lee, Vincent Vanhoucke,
  et~al.
\newblock Socratic models: Composing zero-shot multimodal reasoning with
  language.
\newblock \emph{arXiv preprint arXiv:2204.00598}, 2022.

\bibitem[Zhao et~al.(2020)Zhao, Wu, and Zhang]{MemCap:DBLP:conf/aaai/ZhaoWZ20}
Wentian Zhao, Xinxiao Wu, and Xiaoxun Zhang.
\newblock Memcap: Memorizing style knowledge for image captioning.
\newblock In \emph{The Thirty-Fourth {AAAI} Conference on Artificial
  Intelligence, {AAAI} 2020, The Thirty-Second Innovative Applications of
  Artificial Intelligence Conference, {IAAI} 2020, The Tenth {AAAI} Symposium
  on Educational Advances in Artificial Intelligence, {EAAI} 2020, New York,
  NY, USA, February 7-12, 2020}, pp.\  12984--12992. {AAAI} Press, 2020.
\newblock URL \url{https://ojs.aaai.org/index.php/AAAI/article/view/6998}.

\bibitem[Zhu et~al.(2022)Zhu, Zhu, Li, Wu, Li, Wang, and Dai]{Zhu_2022_CVPR}
Xizhou Zhu, Jinguo Zhu, Hao Li, Xiaoshi Wu, Hongsheng Li, Xiaohua Wang, and
  Jifeng Dai.
\newblock Uni-perceiver: Pre-training unified architecture for generic
  perception for zero-shot and few-shot tasks.
\newblock In \emph{Proceedings of the IEEE/CVF Conference on Computer Vision
  and Pattern Recognition (CVPR)}, pp.\  16804--16815, June 2022.

\end{thebibliography}
\bibliographystyle{iclr2024_conference}

\appendix
\section{Appendix}
\subsection{Datasets}
We split Senticap into train, validation, and test subsets with a ratio of 0.57, 0.13, and 0.3 respectively.
We ended up with a train set of 1,217 images, validation set of 265 images and test set listing 743 images.

In Addition, for the Flickrstyle10k dataset, similar to the approach in \cite{capdec:nukrai2022text} we used a split ratio of 0.75, 0.08, and 0.17, resulting in 4409 images for the training set, with 490 and 1000 images allocated for the validation and test sets, respectively.

\subsection{Hyper-parameters}
\label{sec:appendix-hparams}
For the ZeroCap approach we used the original hyper-parameters suggested by  \cite{tewel2021zero}. The image arithmetic with the emojis worked best for us when multiplying the emoji embedding by 0.5.
During all experiment with ZeroCap we applied 5 optimization steps to GPT-2, and searched over 5 beams as suggested by \cite{tewel2021zero}.
For our {\scshape Apollo-Cap}-based models, we adopted the hyper-parameter values from ZeroCap except of the loss weights, $\lambda$, which were tuned on 20 randomly selected images from the validation set of the target benchmark, SentiCap or Flickrstyle10k. In addition, the gradient descent step size $\alpha$ as well as the style softmax temperature were tuned on the same 20 selected images used for tuning the loss weight.
We tuned hyper-parameters separately for each style to maximize the harmonic average of the TIC, style, and fluency metrics detailed in Section \ref{ssec:evaluation-metrics}.
For demonstrating decentralization, we implemented Algorithm \ref{alg:optimize_CLIP_K_V} with gradient step size $\alpha=0.3$.
Table \ref{tab:appolocap-hparams} details the hyper-parameters used in our experiments.

\subsection{DeepMoji}
\label{sec:appendix-deepmoji}
DeepMoji \citep{felbo2017using} is pre-trained model that was trained on millions of paired tweets-emojis in order to assess the emotion of a text. It generates a vector probability $\mathbf{r}\in \mathbb{R}^{64}$. Each index 'i' represents the probability of emoji 'i' representing the text.
The relevant emojis that this model predicts are shown below:\\
\includegraphics[width=\textwidth / 2]{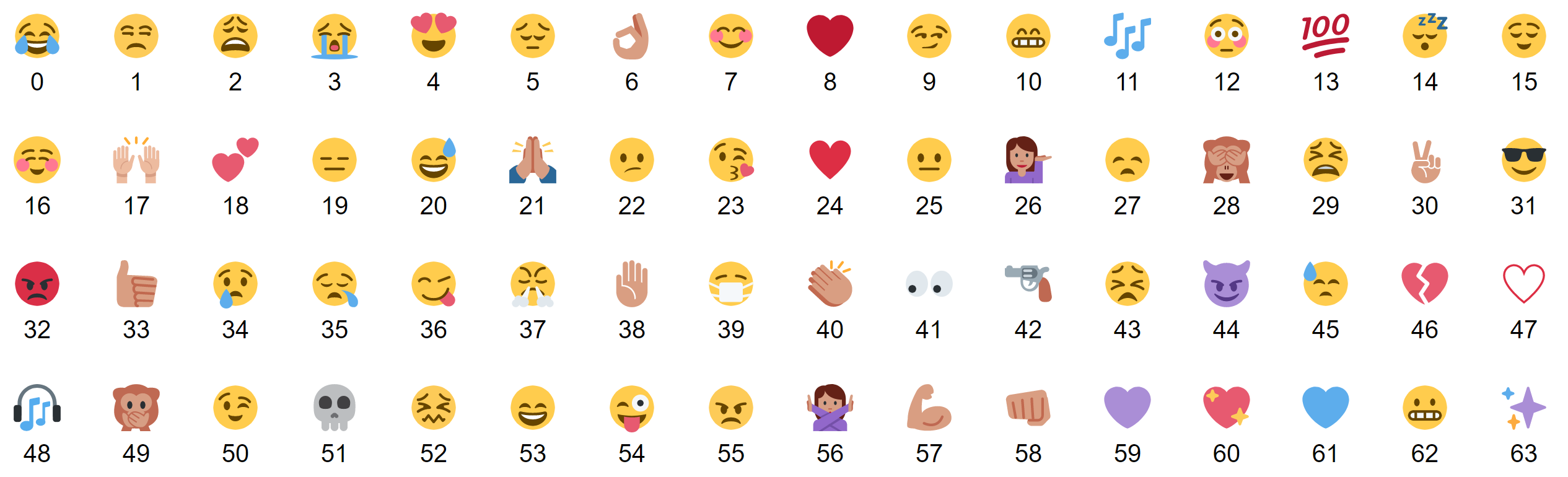}
\\To represent a humorous style, we summed the probabilities associated with emoji indices \{0, 53\}. Conversely, for a romantic style, we focused on emojis with indices \{4, 8, 18, 23, 24\}.

\begin{table}
\centering
\begin{tabular}{l|l|l|l}
\hline
\textbf{Approach} & \textbf{Param} & \textbf{Pos} & \textbf{Neg} \\ \hline
\multirow{3}{*}{\textbf{\scshape Apollo-Cap-PD}} & $\tau$ & 0.14 & 0.17 \\ 
& $J$ & 1 & 1 \\ 
& $\lambda_{LM}$ & 0.22 & 0.61 \\ 
& $\lambda_{CL}$ & 1 & 2 \\ \hline
\multirow{3}{*}{\textbf{\scshape Apollo-Cap-P}} & $\tau$ & 0.01 & 0.09 \\ 
& $\lambda_{LM}$ & 4 & 0.62 \\ 
& $\lambda_{CL}$ & 8 & 2 \\ \hline 
\multirow{4}{*}{\textbf{\scshape Apollo-Cap}} & $\tau$ & 0.001 & 0.001 \\ 
& $\lambda_{CL}$ & 2.2 & 5 \\ 
& $\lambda_{SL}$ & 9.7 & 11.9 \\ 
& $\lambda_{LM}$ & 2 & 2.9 \\ \hline
\end{tabular}
\caption{\textsc{Apollo-Cap} hyper-parameters used for SentiCap}
\label{tab:appolocap-hparams}
\end{table}

\subsection{Qualitative Results for Audio-Aware Image Captioning}
\begin{figure}[h]
\centering
\scalebox{0.8}{
\includegraphics[width=\textwidth]{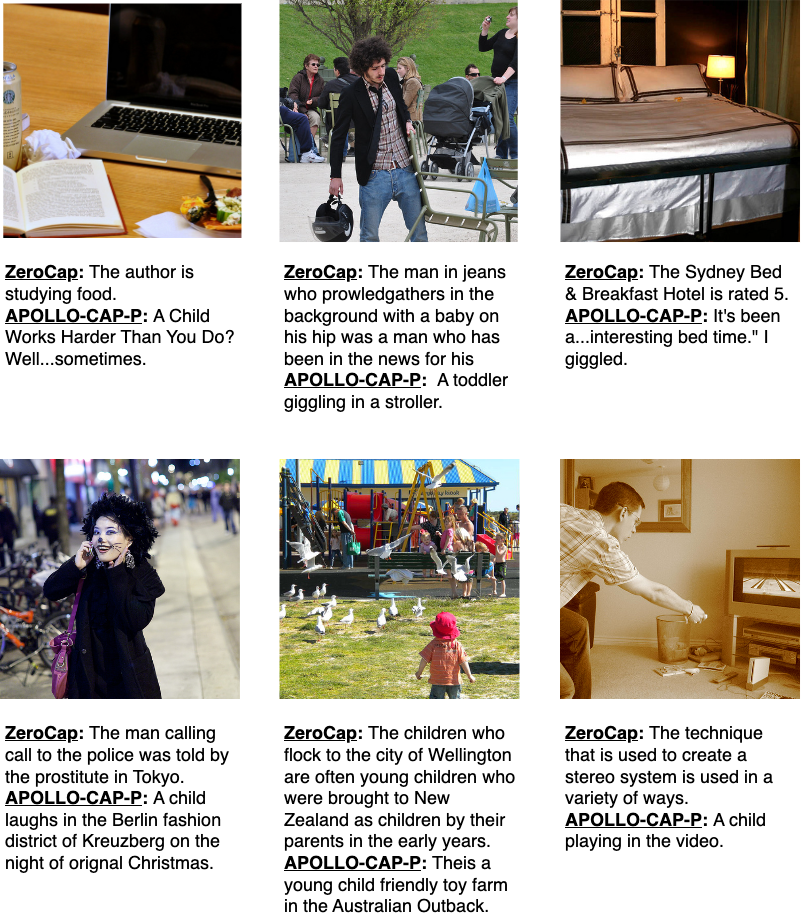}}
 \caption{{\scshape Apollo-Cap-P} caption examples for images and audio clips featuring children's laughter.
 } 
\label{fig:apollo-cap-clap-appendix}
\end{figure}

Figure \ref{fig:apollo-cap-clap-appendix} showcases supplementary results obtained using our model, {\scshape Apollo-Cap-P}, on six test images from the SentiCap dataset. Each image is accompanied by audio featuring children's laughter. To underscore the impact of the audio, we have included ZeroCap's results, which were generated for the images without audio.

\end{document}